\documentclass{article}
\usepackage[preprint,nonatbib]{neurips_2024}
\usepackage[pdftex]{graphicx}
\usepackage{tcolorbox}

\usepackage[utf8]{inputenc} %
\usepackage[T1]{fontenc}    %
\usepackage{hyperref}       %
\usepackage{url}            %
\usepackage{booktabs}       %
\usepackage{amsfonts}       %
\usepackage{nicefrac}       %
\usepackage{microtype}      %
\usepackage{xcolor}         %
\usepackage{listings}
\usepackage{graphicx}

\definecolor{codegreen}{rgb}{0,0.6,0}
\definecolor{codegray}{rgb}{0.5,0.5,0.5}
\definecolor{codepurple}{rgb}{0.58,0,0.82}
\definecolor{backcolour}{rgb}{0.95,0.95,0.92}

\lstdefinestyle{mystyle}{
    backgroundcolor=\color{backcolour},   
    commentstyle=\color{codegreen},
    keywordstyle=\color{magenta},
    numberstyle=\tiny\color{codegray},
    stringstyle=\color{codepurple},
    basicstyle=\ttfamily\footnotesize,
    breakatwhitespace=false,         
    breaklines=true,                 
    captionpos=b,                    
    keepspaces=true,                 
    numbers=left,                    
    numbersep=5pt,                  
    showspaces=false,                
    showstringspaces=false,
    showtabs=false,                  
    tabsize=2
}

\newcommand{\todo}[1]{}
\renewcommand{\todo}[1]{{\color{red} TODO: {#1}}}

\newcommand{\note}[1]{}
\renewcommand{\note}[1]{{\color{blue} TODO: {#1}}}

\title{VQA Training Sets are Self-play Environments for Generating Few-shot Pools}

\author{%
  Tautvydas Misiunas \\
  Google DeepMind\\
  \texttt{tautis@google.com} \\
  \And
  Hassan Mansoor \\
  Google DeepMind\\
  \texttt{hassan@google.com} \\
  \And
  Jasper Uijlings \\
  Google DeepMind\\
  \texttt{jrru@google.com} \\
  \And
  Oriana Riva \\
  Google DeepMind\\
  \texttt{oriva@google.com} \\
  \And
  Victor Carbune \\
  Google DeepMind\\
  \texttt{vcarbune@google.com} \\
}

\begin{document}

\maketitle

\begin{abstract}
Large-language models and large-vision models are increasingly capable of solving compositional reasoning tasks, as measured by breakthroughs in visual-question answering benchmarks. However, state-of-the-art solutions often involve careful construction of large pre-training and fine-tuning datasets, which can be expensive. The use of external tools, whether other ML models, search engines, or APIs, can significantly improve performance by breaking down high-level reasoning questions into sub-questions that are answerable by individual tools, but this approach has similar dataset construction costs to teach fine-tuned models how to use the available tools. We propose a technique in which existing training sets can be directly used for constructing computational environments with task metrics as rewards. This enables a model to autonomously teach itself to use itself or another model as a tool. By doing so, we augment training sets by integrating external signals. The proposed method starts with zero-shot prompts and iteratively refines them by selecting few-shot examples that maximize the task metric on the training set. Our experiments showcase how Gemini learns how to use itself, or another smaller and specialized model such as ScreenAI, to iteratively improve performance on training sets. Our approach successfully generalizes and improves upon zero-shot performance on charts, infographics, and document visual question-answering datasets.
\end{abstract}

\section{Introduction}

 Visual reasoning is a key capability for intelligent agents that perform complex tasks on behalf of users. Challenges range from simple visual information lookup, followed by compositional reasoning on real-world images~\cite{chen2023palix} or on visual representations, such as graphical user interfaces, infographics and charts~\cite{baechler2024screenai, you2024ferretui}. Visual question-answering (VQA) benchmarks capture some of this complexity through rich images and complex visual reasoning questions \cite{zellers2019recognition,mathew2021infographicvqa}. Most recent breakthroughs have been obtained through pre-training and fine-tuning using carefully constructed data mixtures and scalable model architectures \cite{geminiteam2023gemini, openai2024gpt4, mckinzie2024mm1}. We posit that the benefits for downstream users of these models stem more from the ability and flexibility to perform visual in-context learning on any given task, rather than the performance on the specific downstream tasks. Not only the task of interest may not be present in the data mixture, but the various training stages may inadvertently degrade performance because of the large number of tasks involved. Therefore, an emerging class of approaches leverages visual in-context learning (ICL) capabilities \cite{alayrac2022flamingo} for solving VQA tasks without modifying the base model; an LLM orchestrates tools \cite{hu2023avis}, writes code \cite{suris2023vipergpt, stanic2024truly, Gupta_2023_CVPR} or a mix of both \cite{castrejon2024hammr,yang2023mmreact,yao2023react}. Past certain scale \cite{mckinzie2024mm1}, ICL combines modalities (image) with capabilities (code generation) leading to powerful problem solving mechanisms.

\begin{figure}
\label{fig:training_loop}
  \centering
  \includegraphics[trim={0cm 3.5cm 0cm 0cm},width=1.0\linewidth,clip]{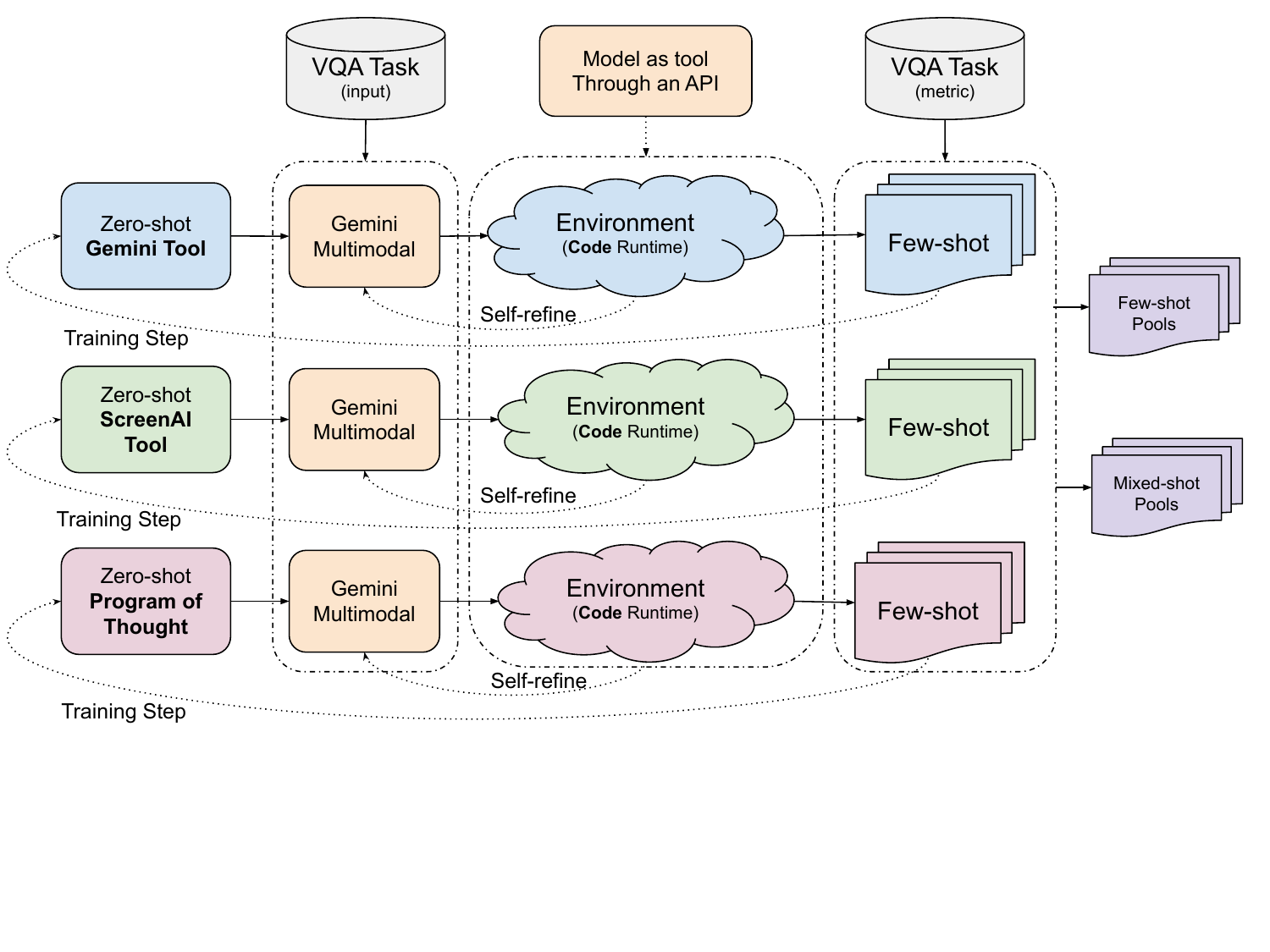}
  \caption{Training sets are transformed into self-play environments where Gemini learns how to use itself, or another model such as ScreenAI, as a tool. Each training step uses the VQA task metric as reward by filtering correctly solved examples and using them as few-shot example in the next round. We seed the environments with zero-shot prompts using three different code generation approaches.}
\end{figure}

Our work focuses on enhancing multimodal reasoning capabilities of vision-language models (VLMs) on difficult reasoning tasks that involve visual representations through a novel few-shot regime not previously studied. We focus our work on most difficult benchmarks available today for charts, infographics and document reasoning: ChartQA~\cite{masry2022chartqa}, PlotQA v2~\cite{methani2020plotqa}, InfographicVQA~\cite{mathew2021infographicvqa}, and DocVQA~\cite{mathew2021docvqa}. Taking inspiration from VisProg~\cite{Gupta_2023_CVPR} and ViperGPT~\cite{suris2023vipergpt}, we generate code conditioned on images using APIs and execute the generated code. We leverage a self-refinement prompt \cite{madaan2023selfrefine} that incorporates execution errors and we compare the result with the golden label in the training set~\cite{stanic2024truly}. Our key insight is to iteratively repeat the process and replace the zero-shot prompt with few-shots that matched the training labels. We seed the process with multiple initial zero-shot prompts which differ through the type of code generated (e.g., program of thought \cite{chen2023program}, API-based \cite{patil2023gorilla}). Our method, visualized in Figure~\ref{fig:training_loop}, treats \emph{training sets as environments}. It iteratively expands the few-shot examples thus constructing several few-shot pools (for each initial zero-shot prompt) or mixed-shot pools (combining across types of zero-shot prompts). These pools are then used at inference time. Unlike prior work~\cite{Gupta_2023_CVPR,suris2023vipergpt} this process does not require any human involvement. For powerful VLMs, such as Gemini \cite{geminiteam2023gemini}, which both generate code and perform visual reasoning, the training loop resembles self-play \cite{silver2017mastering}, where the model learns how to best use itself to solve a VQA task. Furthermore, it taps into improved reasoning capabilities of models which were also trained on code generation datasets \cite{ma2023training}.

\begin{figure}
\label{fig:code_image}
  \centering
  \includegraphics[width=1.0\linewidth]{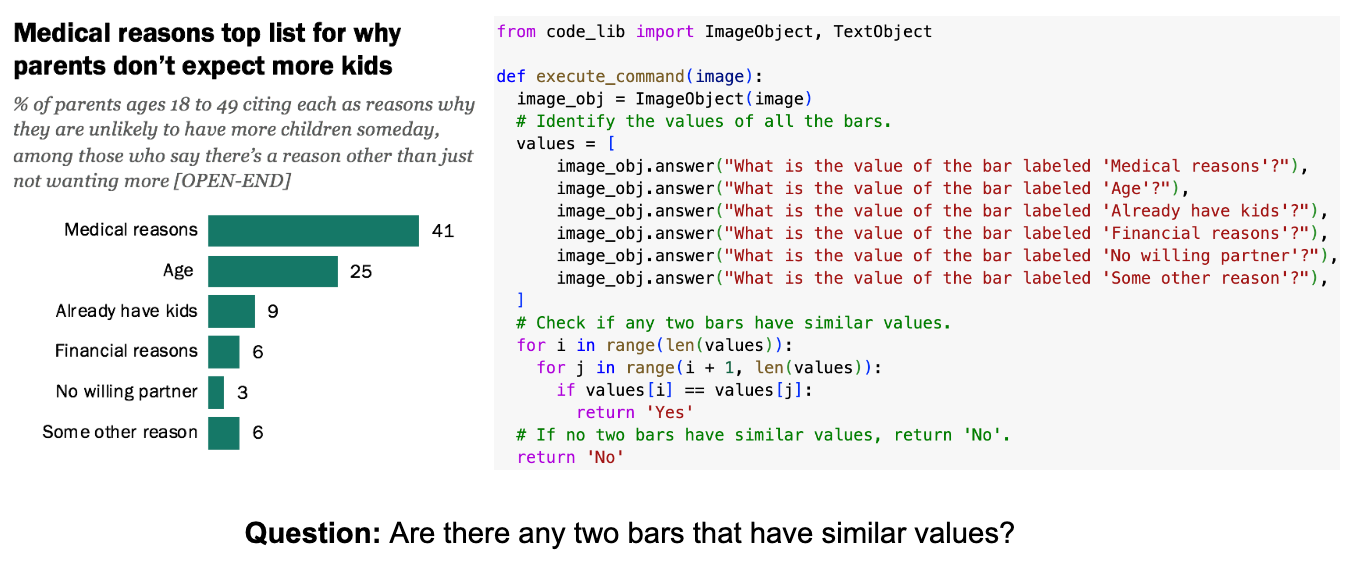}
  \caption{Example of a compositional reasoning question from ChartQA \cite{masry2022chartqa}. Gemini predicts code conditioned on the image, re-using itself through an API for visual information lookup (\texttt{image\_obj.answer}) and leveraging the computational environment for the arithmetic comparison (i.e., comparing bar values).}
\end{figure}

The computational environment constructed through code generation couples values inferred directly from the image (e.g., values of bars or text labels in a chart) with tool (model)'s inference outputs, through basic arithmetic computations \cite{patil2023gorilla,schick2023toolformer,cai2024large}. A main challenge to address in this setup is how to automatically choose APIs to be used in prompts for invoking individual models (as tools). Prior work has shown how LLMs are capable to perform self-debugging and self-correction~\cite{stanic2024truly}. 
We study two types of computation environments, through zero-shot descriptions of them. The first type extends program-of-thought~\cite{bi2023programofthoughts} conditioning on multimodal input, while the second type makes use of an indirection API, enabling the model to focus on orchestration and decomposition aspects.
The self-play environment surfaces and maintains as few-shot those training examples where the model successfully performed such orchestration. Furthermore, it enables a model such as Gemini, to be both an orchestrator, by predicting the code which decomposes the problem and leverages tools, and a tool, by being called by the underlying API. By keeping the training step outputs for which results match training labels, we form few-shot pools for each type of zero-shot prompt provided. We provide an example question and model generated solution using our work in Figure~\ref{fig:code_image}.

Our contributions can be summarized as follows \textbf{(i)} we introduce a novel framework for constructing self-play environments for VLMs, such as Gemini, reusing existing training sets, \textbf{(ii)} we report for the first time performance on PlotQA v2 in few-shot regime and strong performance on DocVQA, InfographicVQA and ChartQA compared to their zero-shot baselines, and \textbf{(iii)} we report experimental results quantifying performance of the individual components of our designed self-play training loop.

\section{Related Work}
\label{sec:related_work}

Strong capabilities of recent multimodal models would hint that specific modalities (e.g., audio, image or video) may become a mere task detail in a zero-shot or few-shot prompt. %
However, it is expected that capabilities may differ by modality, due to specific technical challenges stemming from modality-specific tokenization \cite{borsos2023audiolm,fu2022violet,dosovitskiy2021image} and availability of mixed-modality pre-training datasets \cite{mckinzie2024mm1,fu2022violet} to learn inter-modal dependencies. While impressive results are reported for text modality in many-shot regime \cite{agarwal2024manyshot}, earlier few-shot results on images flattened more quickly \cite{alayrac2022flamingo} and more recent work \cite{jiang2024manyshot} focused on classification tasks highlighted scaling difficulties. Our approach would directly be accelerated by further breakthroughs in multimodal many-shot regime.

\paragraph{Learning to use tools} Toolformer \cite{schick2023toolformer} introduced a pre-training and fine-tuning recipe for augmenting LLMs with capabilities to use tools. ReAct \cite{yao2023react} leverages few-shot capabilities, and has recently been extended \cite{yang2023mmreact, castrejon2024hammr, hu2023avis, gao2023assistgpt} to the multimodal domain. Multi-agent  frameworks such as AutoGen \cite{wu2023autogen} are examples of mainly natural language based environments for learning how to collaborative solve tasks. There, code generation is mainly a tool, however as an environment it can also be very effective at scaling tool use to thousand of APIs \cite{patil2023gorilla}, with selecting among prompt libraries for using specific tools depending on the task being a key element \cite{paranjape2023art}.  Our proposed technique treats predicted code as environments where agents learn to use themselves, a somewhat less explored angle \cite{stanic2024truly, suris2023vipergpt, subramanian-etal-2023-modular}.

\paragraph{Visual QA} Solving visual question-answering tasks poses numerous challenges for VLMs. It requires improved general image representations \cite{alayrac2022flamingo,chen2023palix,baechler2024screenai}, eventually conditioned on questions \cite{ganz2024question,yang2024improving} or even highly specialized on the type of task \cite{carbune2024chartbased,chen2024internvl,levy2022classificationregression}. Such methods require numerous pre-training and fine-tuning experiments and are less flexible compared to those that leverage in-context learning \cite{brown2020language} for improving task performance either through few exemplars \cite{alayrac2022flamingo,song-etal-2022-clip}, or zero-shot techniques such as chain-of-thought \cite{wei2023chainofthought}. Our work leverages both zero-shot capabilities, as well as few-shot learning in a way that enables an iterative refinement loop not previously explored for these tasks.

\paragraph{Self-play Environments} AlphaGo \cite{silver2017mastering} and, more broadly, game environments such as Atari \cite{mnih2013playing} have already been extensively used for training models using reinforcement learning (RL) \cite{tdgammon95}. Game environments not only have a predefined reward function, but the enormous state space cannot be easily captured in offline datasets either. We pose the challenge of solving a VQA task as a self-play environment, where Gemini first constructs an artificial program (the self-play environment) which decomposes the question into multiple steps and then uses itself for answering the underlying simpler questions, ultimately the reward being whether the answer was reached or not. We hypothesize that transforming training datasets this way, paired with richness of programs generated enable VLMs to construct a rich state-space from which compositional reasoning can be improved.

\section{Method}
\label{sec:method}

\begin{figure}
\label{fig:vpt_image}
  \centering
  \includegraphics[trim={0cm 6.5cm 4cm 2.0cm},width=1.0\linewidth,clip]{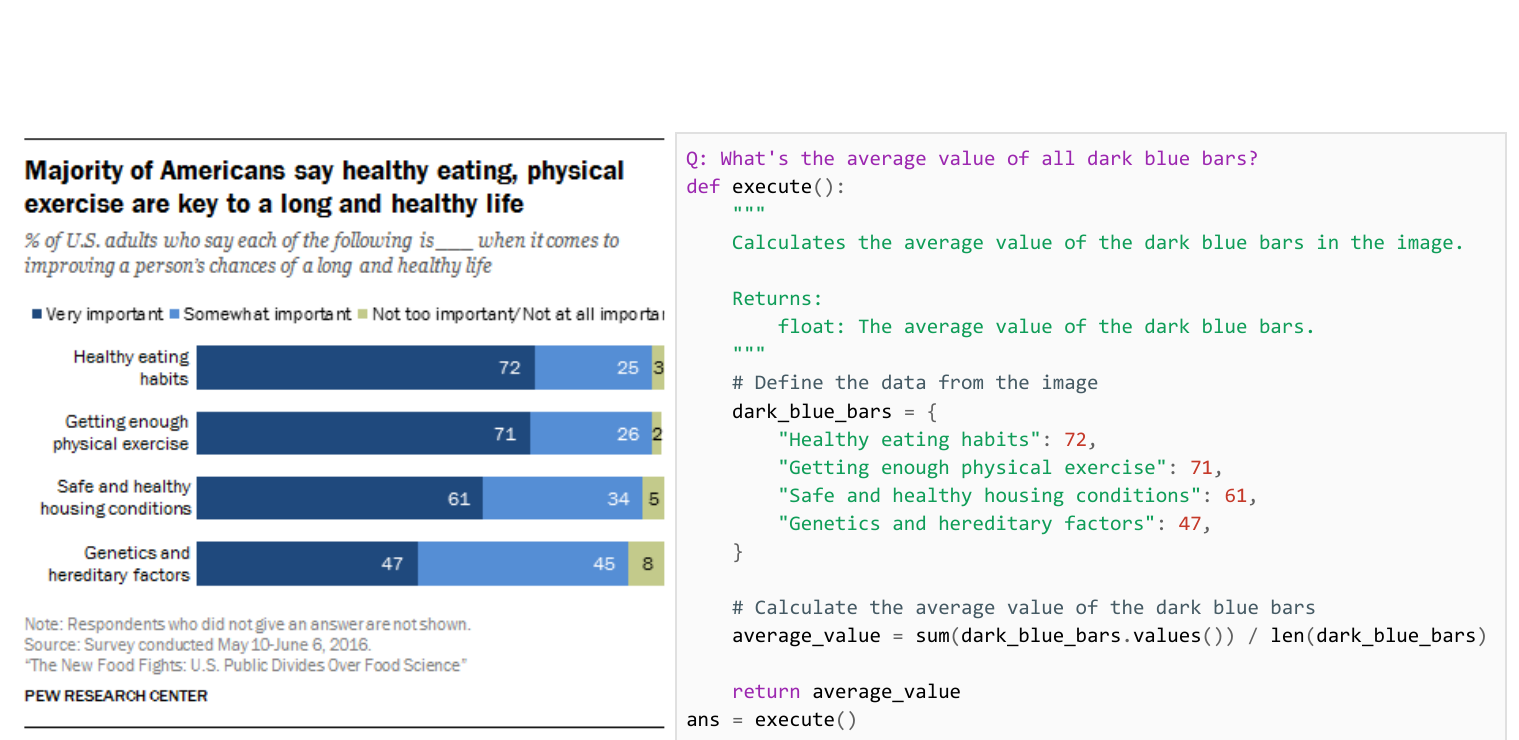}
  \caption{Visual program-of-thought \cite{chen2023program} creates intermediate data structures using extracted values from the image in order to provide an answer that requires arithmetic computations.}
\end{figure}

Key to our approach presented in Figure~\ref{fig:training_loop} is the construction of a synthetic environment through code generation which is able to iteratively improve the performance on the task of interest. We do so by leveraging the zero-shot capabilities of Gemini to describe at a high-level what those environments should be. It is at this stage that instruction-following capabilities, as well as minimal self-refinement capabilities are necessary in order to bootstrap the initial few-shot generation by correctly solving a few examples. Our method can be described through the following stages, a training step with refinement which constructs the few-shot pools for a task, followed by an inference step to sample from the few-shot pool a few exemplars to solve the task.

\paragraph{Training step} Once a few examples are solved using zero-shot, they become few-shot examples. These are used on the training sets of the VQA task of interest for generating additional examples. The ideal pool of few-shots that is constructed per zero-shot prompts would contain sufficiently diverse exemplars to cover a larger task distribution. We detail the zero-shot prompts in Section~\ref{sec:vpt} and Section~\ref{sec:vlmt} and treat the number of training steps as a hyperparameter. 

\paragraph{Refinement step} Given the importance of the initial exemplars, a self-refinement step fixes obvious coding mistakes - such as importing non-existing modules, performing arithmetic operations on strings, etc. We sample multiple times the initial code for diversity in the generated code.

We control the type of environment constructed through the zero-shot prompts. We choose to focus on computational environments which perform visual lookup of information, paired with arithmetic operations and construction of non-trivial data structures and code flows to infuse additional learning signal in existing training sets. Most of our tasks are compositional reasoning questions, oftentimes looking up and comparing quantities, for which such setups are rather well suited.

\subsection{Visual program-of-thought}
\label{sec:vpt}

Our first type of environment is a natural extension to program-of-thought \cite{chen2023program}, where the image alongside the question is used when generating programs, with code interleaved with rationales as comments. Values on the images are extracted directly in code, but there is no API that enables a tool call. An illustration of the generated code is in Figure~\ref{fig:vpt_image} and the corresponding zero-prompt is in the Appendix~\ref{app:vpt}.

\subsection{Vision-language models as tools}
\label{sec:vlmt}

We extend Visual Program-of-Thought with a simple, yet powerful indirection API. It consists of an $ImageObject$ that wraps a provided image and has a semantic $answer$ method for querying a configured model to answer a question. The same API interface can be implemented in different ways and abstracts away details such as which model is called, how it is called and even what hyperparameters are used. The interface is depicted below.

\begin{lstlisting}[basicstyle=\small, language=Python]
class ImageObject:
  """Image class which holds the image and is able to answer questions."""
  def __init__(self, image: Image):
    pass
  def answer(self, question: str) -> str:
  """Method that answers a question by inspecting the image it holds."""
    pass
  def describe(self) -> str:
  """Method that describes the image it holds."""
    pass
\end{lstlisting}

We use two models behind the implementation of the API. Through the first implementation of the API, we set up Gemini as a tool by calling it with a small variation to the standard zero-shot prompt for answering a question on an image, detailed in the Gemini technical report \cite{geminiteam2023gemini}. As a second implementation of the API, we use ScreenAI \cite{baechler2024screenai}, which is a unified 5B parameter VLM for understanding UI interfaces, infographics and charts. Pre-training and fine-tuning mixtures consist of large-scale in-domain image and text interleaved web data, screen annotations, question answering and many more. ScreenAI is expected to provide reasonable performance when used within the code runtime through the API interface.

\section{Experimental Setup and Results}
\label{sec:experimental_setup}

\subsection{VQA Tasks}
\label{sec:vqa_tasks}

We study the effectiveness of our methods on visual-question answering tasks that require compositional reasoning, specifically on images that depict (rich) visual representations such as charts, infographics and documents. For charts, we report the performance on ChartQA \cite{masry2022chartqa}, a commonly studied benchmark for VLMs, and PlotQA v2 \cite{methani2020plotqa}, a particularly difficult benchmark of compositional reasoning questions on scientific plots. DocVQA \cite{mathew2021docvqa} quantifies performance of our approach on documents, however questions are somewhat more direct in nature (e.g., visual look-up). InfographicVQA \cite{mathew2021infographicvqa} comes with high variations in image resolutions, making our approach particularly interesting as models behind tools have to solve only the simpler questions on the image.

\paragraph{Metrics} For ChartQA and PlotQA we report the commonly used relaxed accuracy (RA) metric, where string answers are matched exactly and numeric answers are matched with a 5\% tolerance to ground truth labels. For InfographicVQA and DocVQA the metric used is Averaged Normalized Levenshtein Distance (ANLS), which quantifies the distance from the (multiple) ground truth labels. We report the code pass rate (CPR) as well, as our methods solve these tasks through generated code.

\paragraph{Models} We use Gemini 1.5 Pro exposed in Cloud APIs, the version prior to May 2024 \cite{geminiteam2024may}. For ScreenAI, we make use of a generalist variant described in the associated paper \cite{baechler2024screenai}.

\paragraph{Datasets} Our setups make use of only 1000 examples, instead of the full training and validation sets. We evaluate on the full test sets, where openly available. Although we sample only a fraction of those for our method's training loop, having a larger pool simplifies the bootstrapping process, increasing the likelihood of more diverse examples being included in the refinement process. More details in Appendix~\ref{app:dataset_sizes}.

\subsection{Constructing Few-shot Pools on Training Sets}

We first evaluate the performance of zero-shot prompts on individual training sets. The zero-shot prompts contain instructions provided to Gemini for generating the code using the preferred approach. Either visual program-of-thought, or using a simple high-level API interface illustrated in Figure~\ref{fig:code_image} that underneath leverages Gemini or ScreenAI as tools. The zero-shot prompts are provided in Appendix~\ref{app:api_impl}. We report results in Table~\ref{tab:training_zeroshot}.

\begin{table*}[ht]
    \centering
    \begin{tabular}{l|c|c|c|c}\toprule
    No Training & ChartQA Human & PlotQA v2 & InfographicVQA & DocVQA \\
     & \tiny RA\% [CPR\%] & \tiny RA [CPR\%] & \tiny ANLS [CPR\%] & \tiny ANLS [CPR\%] \\
    \midrule
    0-shot [PoT] & 34.3 [68.0] & 9.6 [69.0] & 39.2 [87.0] & 30.1 [46.0] \\
    0-shot [Tool=ScreenAI] & 39.2 [95.3] & 5.8 [94.1]  & \textbf{43.9} [92.7] & \textbf{79.4} [96.9] \\
    0-shot [Tool=Gemini] & \textbf{44.9} [89.7] & \textbf{13.0} [86.3] & 35.6 [89.2] & 46.2 [92.7] \\
    \bottomrule
    \end{tabular}
    \caption{Performance on the VQA training set achieved by Gemini multimodal using only the instructions provided in the zero-shot prompts, as the first step of the training loop. Instructions are whether to use program-of-thought directly or ScreenAI or Gemini as tools through an API.}
    \label{tab:training_zeroshot}
\end{table*}

Gemini can more effectively use models through APIs from zero-shot instructions rather than solving the task directly using visual program-of-thought. The use of an API allows Gemini to break down the question in easier questions and delegate those to itself or to ScreenAI. As tools, ScreenAI is better suited for working with infographics and documents, whereas Gemini is better suited for chart reasoning tasks. Intuitively, we hypothesize that for tasks with varying image resolution, this approach combines the strengths of Gemini as a higher-level reasoner and ScreenAI as a specialized perception tool. However, we note that leveraging the zero-shot instructions given in our approach is a rather hard task for models such as Gemini and we have to perform a few self-refinement steps based on the execution output of the predicted code, in order to fix obvious mistakes, such as performing numeric operations with string outputs. The effectiveness of such self-refinement steps is higher when applied during the first training step, rather than on subsequent ones where few-shot prompts replace zero-shot prompts.

Although far from perfect, the zero-shot performance is good enough to provide initial examples that can be used as input for our training technique. We sample from the training examples that were correctly solved and iteratively construct better and better few-shot pools. We report the performance improvements stemming from our training method up to 2 steps in Table~\ref{tab:training_fewshot}.

\begin{table*}[ht]
    \centering
    \begin{tabular}{c|c|c|c|c}
    \toprule
    Training & ChartQA Human & PlotQA v2 & InfographicVQA & DocVQA \\
    {\tiny Tool=Gemini} & \tiny RA [CPR\%] & \tiny RA [CPR\%] & \tiny ANLS [CPR\%] & \tiny ANLS [CPR\%] \\
    \midrule
    0-shot [Step 0] & 44.9 [89.7] & 13.0 [86.3] & 35.6 [89.2] & 46.2 [92.7] \\
    4-shot [Step 1] & 51.6 [89.1] & 14.8 [86.5] & 42.0 [91.0] & 66.3 [99.0] \\
    8-shot [Step 2] & \textbf{60.6} [97.1] & \textbf{15.7} [86.1] & \textbf{49.1} [90.3] & \textbf{76.0} [99.2] \\
    \bottomrule
    \end{tabular}
    \caption{Iterative performance on the VQA training set after refinement steps with Gemini as a tool through an API. Our proposed technique improves with more training steps across all VQA tasks of interest.}
    \label{tab:training_fewshot}
\end{table*}

By iteratively repeating the process, we increase the quality of the few-shot examples. The training set size is roughly 1000 examples and we sample randomly after each step. After the training steps, the examples generated in the last training step form the few-shot examples which we further evaluate on validation and test sets.

\subsection{Main Results on VQA Tasks}

We observe diminishing returns the more steps are done on the training set, and as the few-shot pool increases. The fewer iterations, the less computation is necessary and the more cost-effective our approach is. We already note diminishing returns even with the second training step in Table~\ref{tab:training_fewshot}. We extract the few-shot examples to construct example pools from which to run evaluation on the validation set, using all three methods: program-of-thought directly or ScreenAI or Gemini as tools through an API.

\begin{table}[ht]
    \centering
    \scalebox{0.95}{
    \begin{tabular}{l|c|c|c|c}\toprule
    Few-shot & ChartQA Human & PlotQA v2 & InfographicVQA & DocVQA \\
     & \tiny RA\% [CPR\%] & \tiny RA [CPR\%] & \tiny ANLS [CPR\%] & \tiny ANLS [CPR\%] \\
    \midrule
    0-shot [Step 0, PoT] & 47.2 [79.5] & 8.3 [69.8] & 45.0 [85.7] & 36.6 [56.1] \\
    8-shot [Step 2, PoT] & \textbf{68.0} [98.0] & 17.1 [100.0] & \textbf{63.1} [98.0] & 75.1 [98.0] \\
    \midrule
    0-shot [Step 0, Tool=ScreenAI] & 47.0 [95.9] & 5.7 [98.6] & 36.6 [97.9] & 66.9 [98.2] \\
    8-shot [Step 2, Tool=ScreenAI] & 54.4 [95.6] & 7.7 [98.4] & 41.7 [98.2] & 73.3 [98.8] \\
    \midrule
    0-shot [Step 0, Tool=Gemini] & 51.7 [99.0] & 14.9 [99.6] & 48.0 [98.8] & 58.7 [99.4] \\
    8-shot [Step 2, Tool=Gemini] & 67.4 [96.6] & \textbf{19.0} [100] & 59.5 [97.3] & \textbf{81.8} [100] \\
    \bottomrule
    \end{tabular}
    }
    \vspace{1ex}
    \caption{Performance on the VQA validation sets using 8-shot examples automatically constructed using our approach. We note significant improvements even after just two training iterations, regardless of the method and task, validating the capability to generalize of our approach.}
    \label{tab:validation_fewshot}
\end{table}

\subsubsection{Few-shot Pools} The constructed few-shot examples generalize well, as results reported in Table~\ref{tab:validation_fewshot} confirm. The two training steps do not overfit on particular aspects of the training sets and yield improvements for each method and task combination.

However, using the right tool on the example is likely a better strategy. Therefore, our technique although it initially helps the model teach itself how to use a specific tool, it can also be used to combine the different tools into a single evaluation. Indeed, if we estimate an upper bound by combining the results on the validation set using an Oracle (e.g., always selecting the correct answer, if it exists), we observe that approaches complement each other very well (see Table.~\ref{tab:validation_fewshot_agg}).

Therefore, we further investigate how to best combine the constructed few-shot pools at inference time, instead of aggregating them. The simple n\"aive concatenation of the examples in a single prompt has been shown to be error prone, with the model unable to leverage the examples provided all at once \cite{castrejon2024hammr}.

\subsubsection{Mixed-shot Pools} 

\paragraph{Sampling Strategies} In order to combine few-shot pools, a sampling strategy can be used that can compute a similarity between the test example and the examples available in the few-shot pools. Although this overcomes the limitation of n\"aive concatenation, we have not found statistically significant changes in our results. We experimented with leveraging CLIP-based sampling \cite{hu2023promptcap} for directly identifying matching examples, as well as indirect match through a code complexity cluster. For the cluster construction, we estimate the code complexity \cite{bi2023programofthoughts} and then leverage CLIP-based sampling to map to the right cluster using the image/question pair of the test example. When measuring relaxed accuracy on ChartQA using ScreenAI as a tool, we found that random sampling achieves $54.4\%$, while CLIP-based text-image similarity achieves $54.0\%$. Therefore, we rely on simpler approaches for aggregation.

\begin{table*}[th]
\vspace{-.5cm}
    \centering
    \scalebox{0.95}{
    \begin{tabular}{l|c|c|c|c}\toprule
    Mixed-shot & ChartQA Human & PlotQA v2 & InfographicVQA & DocVQA \\
     & \tiny RA\% [$\Delta\%$] & \tiny RA [$\Delta\%$] & \tiny ANLS [$\Delta\%$] & \tiny ANLS [$\Delta\%$] \\
    \midrule
    \emph{Mixed [Agg=Oracle]} & ~86.8 [+27.6\%] & 34.0 [+78\%] & 84.0 [+18.5\%] & ~~91.3 [+11.6\%] \\
    Mixed [Agg=Majority] & 71.4 [+4.9\%] & 19.3 [+1.6\%] & 69.7 [-1.7\%] & 82.3 [+0.6\%] \\
    Mixed [Agg=VLM-Judge] & \textbf{74.7} [+9.8\%] & \textbf{20.4} [+7.4\%] & \textbf{75.3} [+6.2\%] & \textbf{85.8} [+4.9\%] \\
    \bottomrule
    \end{tabular}
    }
    \caption{VQA task performance on validation sets can be further improved through mixed-shot pools, which aggregate outputs using various strategies from multiple few-shot pools independently. $\Delta\%$ indicates improvement compared to the single best performing run in the mixture reported in Table~\ref{tab:validation_fewshot}.}
    \label{tab:validation_fewshot_agg}
\end{table*}

Given the rather neutral impact of sampling strategies, we instead aggregate the outputs from running the few-shot pools independently. Even when running in parallel the constructed few-shot pools, the aggregation technique of the outputs plays a significant role in improving the quality of the results. We observed that casting a \textit{majority vote} \cite{wang2023selfconsistency}, which aggregates results from running each few-shot pool independently, improves performance. Furthermore, the outputs can transform the original question in a multi-choice question with each answer variant coming from independently running inference with a few-shot pool. Then, as an alternative to majority voting, Gemini can rank answers \cite{zheng2023judging}, at a minor increase in computation cost. This approach, referred to as \textit{VLM-Judge} in what follows, can be achieved using the prompt described in Appendix~\ref{app:vlm-as-j}. The headroom from effective sampling strategies can be computed on the validation sets by using an \textit{Oracle} aggregator, selecting the right answer when present among the outputs of independently running the few-shot pools.

\paragraph{Qualitative Changes} We note a few improvements stemming from our approach that improve the quality of the answers, besides the naturally expected improvement of code pass rate that is captured in Table~\ref{tab:training_fewshot}. The written code benefits from subtle improvements; for example, it uses better sorting of the data structure than originally proposed without our method. As another example, without our work, program-of-thought more frequently relies on capturing the answer from the image in a variable and then simply returning the answer. Both example code changes are illustrated in Appendix~\ref{app:improvement_vpt}.

We use four few-shot pools in the aggregation methods in Table~\ref{tab:validation_fewshot_agg}. Three stem from the independent zero-shot prompts that are the seeds of our environments and refined as illustrated in Table~\ref{tab:training_fewshot}, while another one is the standard few-shot pool from training examples. The aggregation is either a simple majority vote or Gemini used as VLM-Judge. Oracle represents an upper bound of the performance, quantifying performance if correct answer was both present and always chosen when running the aggregation technique.

As reported in Table~\ref{tab:validation_fewshot_agg}, the best performing aggregation method is \textit{VLM-Judge}, yielding performance improvements between 5--10\% over the results in Table~\ref{tab:validation_fewshot}.

\begin{table*}[ht]
    \centering
    \scalebox{0.92}{
    \begin{tabular}{l|c|c|c|c}
    \toprule
    Pool & \multicolumn{3}{c|}{ChartQA {\tiny RA\%}} & PlotQA v2 {\tiny RA\%} \\
    & avg & human & augmented & \\
    \midrule
    SoTA & 87.2 \cite{geminiteam2024may} & - & - & 34.2 \cite{levy2022classificationregression} \\
    \midrule
    0-shot, No training [PoT] & 64.3 & 58.2 & 70.4 & 11.9 \\
    Few-shot Pool, 2-step [PoT] & 74.7 & 69.6 & 79.8 & 15.8 \\
    \midrule
    0-shot, No training [Tool=ScreenAI] & 56.6 & 48.6 & 64.5 & 17.8 \\
    Few-shot Pool, 2-step [Tool=ScreenAI] & 73.6 & 66.2 & 81.0 & 19.9 \\
    \midrule
    0-shot, No training [Tool=Gemini] & 56.7 & 50.1 & 63.3 & 15.5 \\
    Few-shot Pool, 2-step [Tool=Gemini] & 72.3 & 66.2 & 78.4 & 18.0 \\
    \midrule
    Mixed-shot Pool [Agg=Majority] & 78.3 [+4.8\%] & 72.8 [+4.4\%] & 83.8 [+3.5\%] & 18.7 [-6.0\%]  \\
    Mixed-shot Pool [Agg=VLM-Judge] & 79.4 [+6.3\%] & 75.3 [+8.2\%] & 83.5 [+3.1\%] & 20.5 [+3.0\%] \\
    \bottomrule
    \end{tabular}}
    \caption{Performance of our method on the open test-sets for ChartQA and PlotQA v2. Our method improves zero-shot performance for all approaches. SoTA results are provided for reference and leverage entire training sets, while our approach bootstraps the solutions with a fraction of the examples.}
    \label{tab:test_eval_results}
\end{table*}

Our method exhibits strong generalization across the two tasks with open test-sets, showcasing that aggregation techniques are effective over outputs from multiple few-shot pools. Table~\ref{tab:test_eval_results} illustrates our method's results next to current state-of-the-art (SoTA) performance. For both benchmarks, our approach solves the tasks by bootstrapping solutions using code generation. For ChartQA, performance using code generation can be considered out of the training distribution, whereas for Gemini 1.5 Pro  \cite{geminiteam2024may} direct question answering on images is likely a lot more present in datasets. Our experiments use an earlier version of Gemini 1.5 Pro \cite{geminiteam2023gemini}, which achieved only 81.3\% on ChartQA. For PlotQA v2, SoTA is obtained using a highly specialized approach and model for chart understanding \cite{levy2022classificationregression}. Our self-play environment enables Gemini to self-teach how to use itself better and reach performance in only two iteration steps, with a quality gain larger than what traditionally may take multiple epochs of training \cite{chen2023palix}. We discuss qualitative insights into how our self-play environments evolve the few-shot pools after the two training steps in Appendix~\ref{app:improvement_vpt}.

\section{Conclusion}
\label{sec:conclusion}

Our work introduces a novel training recipe through which highly capable models, such as Gemini multimodal, can leverage their joint image understanding and code generation capabilities for bootstrapping improved performance. We validate our approach by seeding environments with zero-shot prompts that solve a given task in two ways, through program-of-thought or through a indirection API that enables Gemini to focus on the high-level reasoning challenge. Our technique iteratively improves performance on visual-question answering training sets, generalizing strongly on validation and test sets after just a few training steps. Improvements over zero-shot baseline are strong across each environment and task combination. Through our aggregation techniques, our mixed-shot pools successfully leverage multiple few-shot pool at inference time.

\section{Limitations}
\label{sec:limitations}

We report several limitations of our work: we note that the evaluation focused on a limited number of environment types (two) and tools, used with a strong emphasis on a powerful indirection API, but potentially overlooking scalability challenges. Our results on test sets are limited to the open ones, while work on the closed test sets of DocVQA and InfographicVQA will be done at a later stage. Although the effects of few-shot learning were effectively leveraged, a deeper investigation into many-shot learning was not undertaken before iterating on training steps to refine the few-shots. Lastly, we think our work extends to natural images; however, our focus has been on UI interfaces, charts, and infographics.

\section{Societal impact}
\label{sec:societal_impact}

Our work proposes a technique through which a training set can be exploited in creating an environment through which alternative solutions can semi-autonomously, through self-play, be constructed by large models to solve the task. Training sets are already heavily used in the pre-training and fine-tuning stages of large model training. Our work reuses these datasets. Due to inherent limitations of the training sets, it is likely that broader societal impact is rather limited compared to what models already achieve today. The particular skill acquired does play a role, in our case for refining generated code as an alternative problem solving path, should have a positive impact where models can bootstrap alternative approaches the downstream user has access to. However, refining and improving generated code does require safety handles, such as sandboxed environments. Our technique may also have an impact on the type of datasets collected, as our method enables constructing datasets synthetically by introducing external learning signals from the computational environment. It may be able to reduce the collection of unnecessary datasets, being able to extract more from the currently available ones already collected.

\section{Acknowledgements}
\label{sec:acknowledgements}

Several collaborators have helped shape this work with insightful feedback and discussions. Thanks to Jingdon Chen and Abhanshu Sharma for their early support and feedback on the project. We thank Srinivas Sunkara and Gilles Baechler for hands-on help on leveraging ScreenAI models. We thank Duc-Hieu Tran and Florian Hartmann for their regular feedback on the work. Gil Fidel and Yael Karov have helped with insightful conversations on planning and reasoning through code generation using LLMs.
We thank Lluis Castrejon, Thomas Mensink, Andre Araujo, and Vitto Ferrari on their deep discussions surrounding LLMs which orchestrate tools.
We thank Ewa Dominowska and Jay Yagnik for supporting this research.

\bibliographystyle{plain}
\bibliography{bibliography}

\newpage
\appendix

\section{Improvement examples after two training steps}
\label{app:improvement_vpt}

\begin{figure}[ht]
\label{fig:vpt_improvement_image}
  \centering
  \includegraphics[trim={1cm 7cm 10cm 3cm},width=0.9\linewidth,clip]{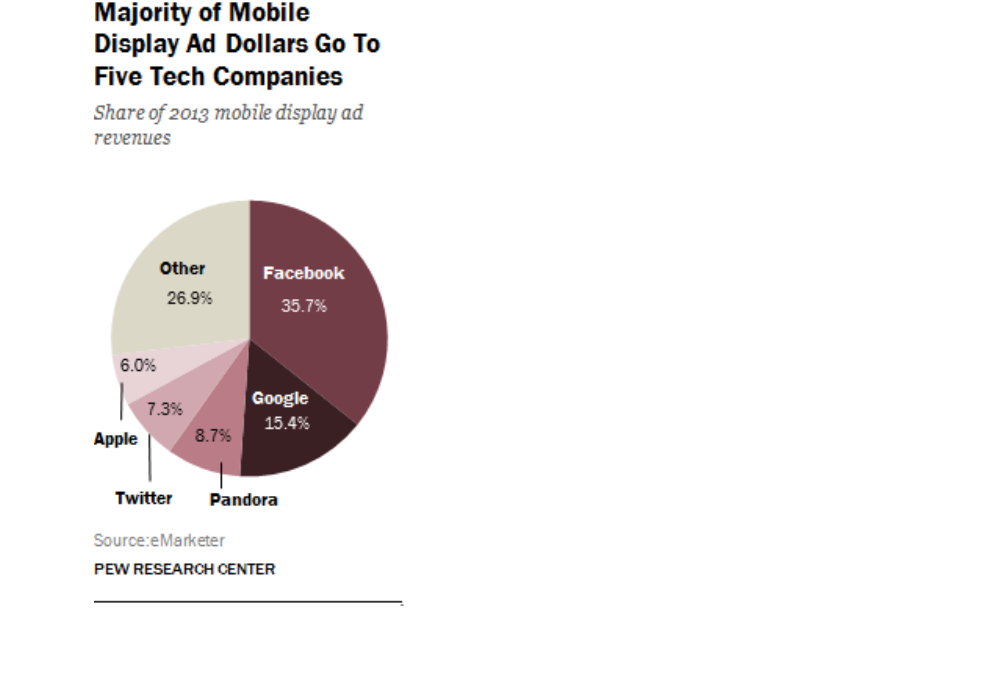}
\end{figure}

Consider the task of identifying the third largest contributor on the following image from the ChartQA dataset in Figure~\ref{fig:vpt_improvement_image}.

The environment seed, our zero-shot prompt for visual program-of-thought, generates the following:
\begin{lstlisting}[language=Python, basicstyle=\ttfamily\footnotesize, style=mystyle]
Q: What is the third largest contributor?
======= Zero-shot (Incorrect) ========
def execute():
  """
  This function returns the third largest contributor in the given pie chart.

  Args:
    None

  Returns:
    A string representing the name of the third largest contributor.
  """
  # Get the data from the pie chart.
  data = [35.7, 15.4, 7.3, 8.7, 26.9]

  # Sort the data in descending order.
  data.sort(reverse=True)

  # Get the third largest value.
  third_largest = data[2]

  # Find the corresponding company name.
  company_names = ['Facebook', 'Google', 'Apple', 'Twitter', 'Other']
  third_largest_company = company_names[data.index(third_largest)]

  # Return the company name.
  return third_largest_company
ans = execute()
\end{lstlisting}

\newpage
After two training steps, where zero-shot examples are iteratively replaced with few-shot examples based on their training set labels, the code is modified as follows:
\begin{lstlisting}[language=Python, basicstyle=\ttfamily\footnotesize, style=mystyle]
Q: What is the third largest contributor?
======== Few-shot Pool (Correct) =======
def execute():
  """
  This function returns the third largest contributor in the graph.

  The answer is Twitter because it has a 7.3%

  Args:
    None

  Returns:
    The third largest contributor in the graph as a string.
  """

  # Get the data from the image.
  data = [
    {
      "company": "Facebook",
      "revenue_share": 35.7
    },
    {
      "company": "Google",
      "revenue_share": 15.4
    },
    {
      "company": "Other",
      "revenue_share": 26.9
    },
    {
      "company": "Apple",
      "revenue_share": 6.0
    },
    {
      "company": "Twitter",
      "revenue_share": 7.3
    },
    {
      "company": "Pandora",
      "revenue_share": 8.7
    }
  ]

  # Sort the data by revenue share.
  sorted_data = sorted(data, key=lambda x: x["revenue_share"], reverse=True)

  # Get the third largest contributor.
  third_largest_contributor = sorted_data[2]["company"]

  # Return the third largest contributor.
  return third_largest_contributor
ans = execute()
\end{lstlisting}

We note that the doc-string where the model tries to solve the question through comments may be an artefact of the training data. Even so, our iterative method fixes the code logic and the correct answer can be returned. 

\begin{lstlisting}[language=Python, basicstyle=\ttfamily\footnotesize, style=mystyle]
Q: What is the value of the largest pie section?
========= Zero-shot(Incorrect) ===========
def execute():
  """
  Computes the value of the largest pie section.

  The largest pie section is the one that represents the largest percentage
  of the pie. In this case, the largest pie section is the one that
  represents the percentage of people who support allowing companies
  from other countries to invest in Pemex. This percentage is 34%

  Args:
    None

  Returns:
    The value of the largest pie section as a float.
  """

  # Get the value of the largest pie section.
  largest_pie_section = 34

  # Return the value of the largest pie section.
  return largest_pie_section
ans = execute()
\end{lstlisting}

The following code is instead predicted after two refinement steps.

\begin{lstlisting}[language=Python, basicstyle=\ttfamily\footnotesize, style=mystyle]
Q: What is the value of the largest pie section?
======== Few-shot Pool (Correct) =======
def execute():
  """Calculates the value of the largest pie section.

  The largest pie section represents the percentage of people who oppose allowing
  companies from other countries to invest in Pemex. The value of the largest pie
  section is 57, which means that 57%
  other countries to invest in Pemex.

  Args:
    None

  Returns:
    The value of the largest pie section.
  """

  # Get the data from the image.
  data = [
      (2007, 41, 44),
      (2009, 43, 44),
      (2011, 57, 49),
      (2013, 51, 37),
      (2015, 23, 19),
      (2017, 41, 29)
  ]

  # Get the U.S. favourability in Russia for each year.
  us_favorability_in_russia = [y[1] for y in data]

  # Get the highest value in U.S favourability in Russia.
  highest_value = max(us_favorability_in_russia)

  return highest_value
ans = execute()
\end{lstlisting}

\section{Zero-shot and Self-refinement Prompts}
\label{app:api_impl}

In this section we provide the zero-shot prompts used for generating the initial examples. Once a few examples are generated on training sets, these zero-shot prompts are replaced by few-shot examples that have successfully matched labels on the training set.

\subsection{Visual Program-of-Thought}
\label{app:vpt}

The first type of prompt, described in Section~\ref{sec:vpt}, is visual program-of-thought.

\begin{tcolorbox}
[width=\linewidth, sharp corners=all, colback=white!95!black]

Look at the image and question pair below. The main objective is to write a function `execute()` to answer the question from the image. \\
In the Python documentation of the function, provide step by step reasoning to explain how the following question can be answered. \\
Afterward write the code that will answer the given question. \\
Return the final answer from the function. All the required information is given in the image. Do not load any external files or request for additional input. \\ 
Pay attention to the units of the answer and when providing percentage as an answer convert the number to decimal format. \\ Write professional level code that an experienced software developer would write. \\
Prefer to write explicit code instead of implicit calculations (e.g. use Python standard libraries to compute max, mean, median values, etc.). \\
Do not print anything with Python print function. Generate Python function only. No english text.
\end{tcolorbox}

\subsection{Gemini and ScreenAI as Tools}

The second type of prompts, involve using Gemini or ScreenAI as tools, as described in Section~\ref{sec:vlmt}. These do not include any implementation detail, e.g. how to call any of the models or what prompts are used when calling them. Instead a generic interface description within a prompt is provided.

\begin{tcolorbox}
[width=\linewidth, sharp corners=all, colback=white!95!black]

Look at the image and question pair below. The main objective is to write a function `execute()` to answer the question from the image. \\
In the Python documentation of the function, provide step by step reasoning to explain how the following question can be answered. \\
Afterward write the code that will answer the given question. \\
Return the final answer from the function. All the required information is given in the image. Do not load any external files or request for additional input. \\ 
Pay attention to the units of the answer and when providing percentage as an answer convert the number to decimal format. \\ Write professional level code that an experienced software developer would write. \\
Prefer to write explicit code instead of implicit calculations (e.g. use Python standard libraries to compute max, mean, median values, etc.). \\
Do not print anything with Python print function. Generate Python function only. No english text. \\ \\
You are given an interface and some examples of how to use the interface to answer the question. Your task to answer a newly given question with the interface. \\

These are interface descriptions of python classes you can use. Actual implementations are provided at runtime. \\

\{INTERFACE\_DESCRIPTION\_PROMPT\}

Here are some examples of what the implementation of it may return: \\
ImageObject(image).answer('What is the value of ...?') may return a number \\ 
ImageObject(image).answer('Is ...?') may return a Yes / No \\
ImageObject(image).answer('What are the steps?') may return a comma-sep string \\

For the execute function make use of the ImageObject class. Only the answer() method. \\

All queries should have an answer, so no need to consider corner cases. \\

For usual cases, follow the guidelines below: \\
- For simple visual queries, directly output the answer in the code. \\
- For queries that require counting and spatial relations, use python code. \\

Consider the following guidelines: \\
- Use base Python (comparison, sorting) for basic logical operations, left/right/up/down, math, etc. \\
- Do not import additional modules and do not use types for variables. \\
- Use only the ImageObject when multiple questions are needed to answer the given question. \\
- When calling answer on ImageObject use as complete and specific questions as possible. \\

The code you output can look similar to this function below \\
\# Question: ... \\
def execute(image): \\
  \# Explanation for why a first step like the one below is needed \\
  im = ImageObject(image) \\
  value = im.answer(question) \\
  \# Explanation for why the next value is needed \\
  other\_value = im.answer(other\_question) \\
  \# Explanation on how to combine the values in a meaningful way for answering the original question. \\
  ans = value + other\_value \\
  return ans \\

Do not print anything with Python print function. \\
Generate Python function only. \\
No english text.
\end{tcolorbox}

\subsection{Self-refinement prompt}

The self-refinement strategy is rather straightforward and is captured through the prompt below.
\begin{tcolorbox}
[width=\linewidth, sharp corners=all, colback=white!95!black]
// Missing: answer variable \\
This code is missing the final answer variable. The final answer should be assigned to the answer variable (\{answer\_var\}). Correct the missing variable mistake and try again. \\
// NameError: usually import statement missing. \\
This code has raised NamedError:  \{error\_trace\}. There might be missing import statements. Correct the NameError mistake and try again. \\
// Generic: for everything else. \\
The code above is a valid Python code, however it raised \{error\_type\}: \{error\_trace\} \\
Correct the mistake and try again please. \\
\end{tcolorbox}

\section{VLM-as-Judge Prompt}
\label{app:vlm-as-j}

\begin{tcolorbox}
[width=\linewidth, sharp corners=all, colback=white!95!black]
Think step by step before giving a final answer to this question. \\
 Format your final answer choice as Final Choice: Answer N: X. \\ \\
For the final answer X, follow the following instructions: \\
** X should be a short and terse answer to the question. \\
** Dont paraphrase or reformat the text you see in the image. \\
** If the final answer has two or more items, provide it in the list format like [1, 2]. \\
** When asked to give a ratio, give out the decimal value like 0.25 instead of 1:4. \\
** When asked to give a percentage, give out the whole value like 17 instead of decimal like 0.17\% \\
** Dont include any units in the answer. \\
** Try to include the full label from the graph when asked about an entity. \\
Additionally you are given four answers from other systems to the same question.  \\
At least on of the given answers is correct. \\
Pick the best answer as final answer from the given choices. \\
If multiple answers are correct, pick the first one. It is important to give only a single correct answer after Final Answer: prefix. \\
Remember, dont give a final answer before step by step reasoning \\
 and format your final answer as Final Choice: Answer N: X. Anything else afterwards will be treated as an incorrect answer. \\
\end{tcolorbox}

\section{Dataset size summary}
\label{app:dataset_sizes}

\begin{table*}[ht]
    \centering
    \scalebox{0.95}{
    \begin{tabular}{l|c|c|c|c|c}\toprule
    Size & ChartQA Human & ChartQA Augmented & PlotQA v2 & InfographicVQA & DocVQA \\
    \midrule
    Training & 1000 & - & 1000 & 1000 & 1000  \\
    Validation & 960 & - & 1000 & 1000 & 1000 \\
    Test & 1250 & 1250 & 1000 & - & -\\
    \bottomrule
    \end{tabular}
    }
    \caption{To reduce costs, we sampled down datasets: all datasets containing 1000 samples were randomly sampled and kept consistent across all runs. We used full sized validation and test sets for ChartQA.}
    \label{tab:dataset_sizes}
\end{table*}

\end{document}